\documentclass[10pt,twocolumn,letterpaper]{article}

\usepackage{cvpr}              

\usepackage{makecell}
\usepackage{multirow}
\usepackage{booktabs}
\usepackage{pifont}
\usepackage[table]{xcolor}
\definecolor{cvprblue}{rgb}{0.21,0.49,0.74}
\usepackage[pagebackref,breaklinks,colorlinks,allcolors=cvprblue]{hyperref}

\definecolor{myblue}{RGB}{30,50,190}
\definecolor{myred}{RGB}{190,40,40}

\definecolor{gaingreen}{rgb}{0.125,0.549,0.129}
\definecolor{losered}{rgb}{0.549,0.125,0.129}
\definecolor{darkgreen}{HTML}{083228}
\newcommand{\sname}{CroBo\xspace}
\newcommand{\snamepp}{CroBo++\xspace}
\newcommand{\gain}[1]{\textbf{{\color{gaingreen}#1}}}
\newcommand{\lose}[1]{\textbf{{\color{losered}#1}}}
\newcommand{\sdag}{\textsuperscript{$\dagger$}}
\newcommand{\dagn}[1]{#1\sdag}

\usepackage{xspace}
\newcommand{\whatiswhere}{\textbf{\textcolor{myred}{what-is-where}}\xspace}
\newcommand{\whatmoveswhere}{\textbf{\textcolor{myblue}{what-moves-where}}\xspace}

\usepackage{cuted}
\usepackage{capt-of}
\usepackage{hyperref}
\usepackage{subcaption}
\usepackage{selectp}

\usepackage{tabularx}
\usepackage{array}

\def\paperID{*****} 
\def\confName{CVPR}
\def\confYear{2026}

\title{
Pixel-level Scene Understanding in One Token: \\
Visual States Need What-is-Where Composition}

\author{
Seokmin Lee \quad Yunghee Lee \quad Byeonghyun Pak \quad Byeongju Woo \\[0.35em]
Agency for Defense Development \\
{\tt\small \{seokminlee, yhl, byeonghyun\_pak\}@add.re.kr \quad byeongju@umich.edu}
}

\begin{document}
\twocolumn[{
\maketitle
}]

\iftrue
    \begin{abstract}
    For robotic agents operating in dynamic environments, learning visual state representations from streaming video observations is essential for sequential decision making.
    Recent self-supervised learning methods have shown strong transferability across vision tasks, but they do not explicitly address what a good visual state should encode.
    We argue that effective visual states must capture \whatiswhere by jointly encoding the semantic identities of scene elements and their spatial locations, enabling reliable detection of subtle dynamics across observations.
    To this end, we propose \textbf{CroBo}, a visual state representation learning framework based on a global-to-local reconstruction objective.
    Given a reference observation compressed into a compact bottleneck token, CroBo learns to reconstruct heavily masked patches in a local target crop from sparse visible cues, using the global bottleneck token as context.
    This learning objective encourages the bottleneck token to encode a fine-grained representation of scene-wide semantic entities, including their identities, spatial locations, and configurations. 
    As a result, the learned visual states reveal how scene elements move and interact over time, supporting sequential decision making.
    We evaluate CroBo on diverse vision-based robot policy learning benchmarks, where it achieves state-of-the-art performance.
    Reconstruction analyses and perceptual straightness experiments further show that the learned representations preserve pixel-level scene composition and encode \whatmoveswhere across observations.
    Project page available at: \href{https://seokminlee-chris.github.io/CroBo-ProjectPage}{https://seokminlee-chris.github.io/CroBo-ProjectPage}.
    \end{abstract}
\fi





\section{Introduction}
\label{sec:intro}

{}

{\iftrue
    The world is inherently dynamic: objects move, agents act, and scene configurations continuously evolve over time.
    For agents to operate reliably in such environments, they must build internal state representations from streams of visual observations and leverage them to support sequential decision making.
    Learning to encode meaningful, task-relevant information from raw visual inputs is therefore a central challenge for real-world video applications, including robot learning and world modeling.
\fi}

\begin{figure}[!t]
    \centering
    \includegraphics[width=0.98\linewidth]{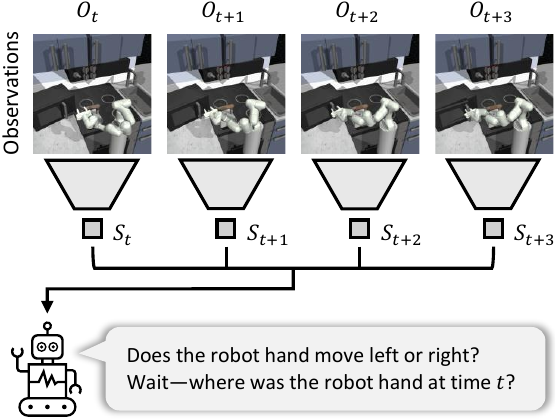}
    \caption{
    {\bf Understanding \whatmoveswhere requires knowing \whatiswhere.}
    Robots interact with dynamic environments through compact state representations derived from visual observations. 
    To reason scene dynamics, the robot must understand which objects move and detect how their locations change between observations. This requires each state to capture \whatiswhere information in the scene so that subtle spatial differences between observations can be recognized.
    }
    \label{fig:concept}
    \vspace{-4mm}
\end{figure}

{}

{\iftrue
    Recent studies~\cite{he2020momentum,chen2020improved,chen2021empirical,he2022masked,caron2021emerging,oquab2023dinov2} have demonstrated that self-supervised learning (SSL) methods yield representations with strong transferability to a wide range of downstream vision tasks, such as image classification and semantic segmentation.
    For robotic agents in particular, however, learning a state representation poses a distinct challenge: the representation must support action by compressing raw observations into a compact visual state while preserving the information essential for decision making~\cite{echchahed2025survey}.
    As a recent step in this direction, ToBo~\cite{kim2025token} introduces a bottleneck token that is trained to reconstruct a heavily masked subsequent frame from a reference observation, using only sparse target patches as hints.
    This formulation encourages the vision encoder to form compact yet temporally aware scene representations.
    While this approach achieves strong performance on robot learning benchmarks, a fundamental question remains insufficiently addressed: \emph{what should a good visual state representation actually encode?}
\fi}

{}

{\iftrue
    We argue that a visual state representation for sequential decision making must capture the \textit{\whatiswhere} scene composition by encoding both the semantic identities of scene elements and their precise spatial locations.
    Here, \whatiswhere denotes whether the representation retains which semantic entities are present in the scene and how they are spatially located and arranged within the overall scene composition.
    For example, as illustrated in \cref{fig:concept}, to detect that a robot hand moved from right to left across observations, the representation must encode both the identity of the hand and its position, so that even subtle spatial changes can be directly detected.
    From this perspective, understanding scene dynamics in robotics can be viewed as a form of pixel-level video understanding, where the representation must preserve spatial semantics while remaining sensitive to how they evolve across observations.
\fi}

\begin{figure}[!t]
    \centering
    \includegraphics[width=1\linewidth]{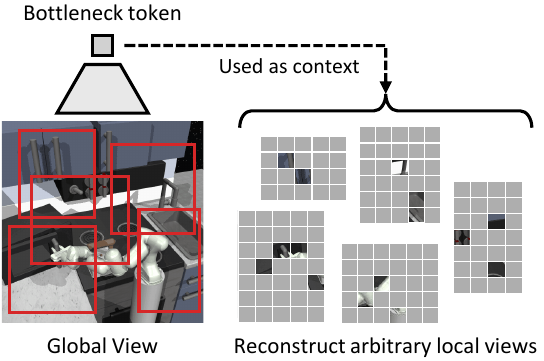}
    \caption{
    {\bf Learning a visual state that captures pixel-level scene composition.}
    We present a state representation learning (SRL) method that captures the entire scene composition, including object identities, locations, and their spatial relationship. 
    %
    Conceptually, the global state becomes the bottleneck token that contains the contextual information of the scene.
    By enforcing reconstruction of arbitrary cropped views from such contextual information, the model is encouraged to encode pixel-level scene composition information in the global state.
    }
    \label{fig:concept2}
    \vspace{-4mm}
\end{figure}

{}

{\iftrue
    Based on this insight, we propose \sname, a simple yet effective state representation learning framework designed to capture \whatiswhere visual states.
    As illustrated in \cref{fig:concept2}, \sname builds upon the bottleneck formulation of ToBo~\cite{kim2025token} and leverages a \emph{global-to-local reconstruction objective} tailored for scene composition learning.
    Given a reference global observation, the model produces a compact bottleneck serving as a contextual memory of the scene, and is trained to reconstruct an arbitrarily and heavily masked local crop using only sparse visible hints from the crop itself.
    To solve this reconstruction task, the model must infer where the crop originates within the scene and what semantic content should appear there, thereby forcing the representation to preserve fine-grained, pixel-level \whatiswhere information across the full observation. 
    In this way, \sname\ learns a compact visual state representation tightly aligned with the requirements of downstream action in dynamic environments.
\fi}

{}

{}

We validate CroBo through an extensive set of experiments.
On vision-based robot policy learning benchmarks, our method achieves state-of-the-art performance.
Beyond policy learning results, qualitative reconstruction analyses show that the learned visual state representation indeed captures pixel-level \whatiswhere in the scene.
We further show through perceptual straightness experiments that this representation better captures \whatmoveswhere across observations.
Finally, ablation studies demonstrate individual gains from different objectives, showing the effectiveness of our global-to-local reconstruction approach.
In summary, our contributions are threefold:
\begin{itemize}
    \item We identify that understanding scene dynamics for robot learning requires visual state representations that encode pixel-level \whatiswhere, rather than relying on temporal or patch-level correspondence.
    \item We introduce CroBo, a self-supervised learning framework that enforces such scene understanding by reconstructing heavily masked local views from a compact representation. 
    \item We show that CroBo achieves state-of-the-art on robot policy learning tasks with representations effectively capturing \whatmoveswhere.
\end{itemize}

\section{Method}
\label{sec:method}

\begin{figure*}[!t]
    \centering
    \includegraphics[width=0.93\linewidth]{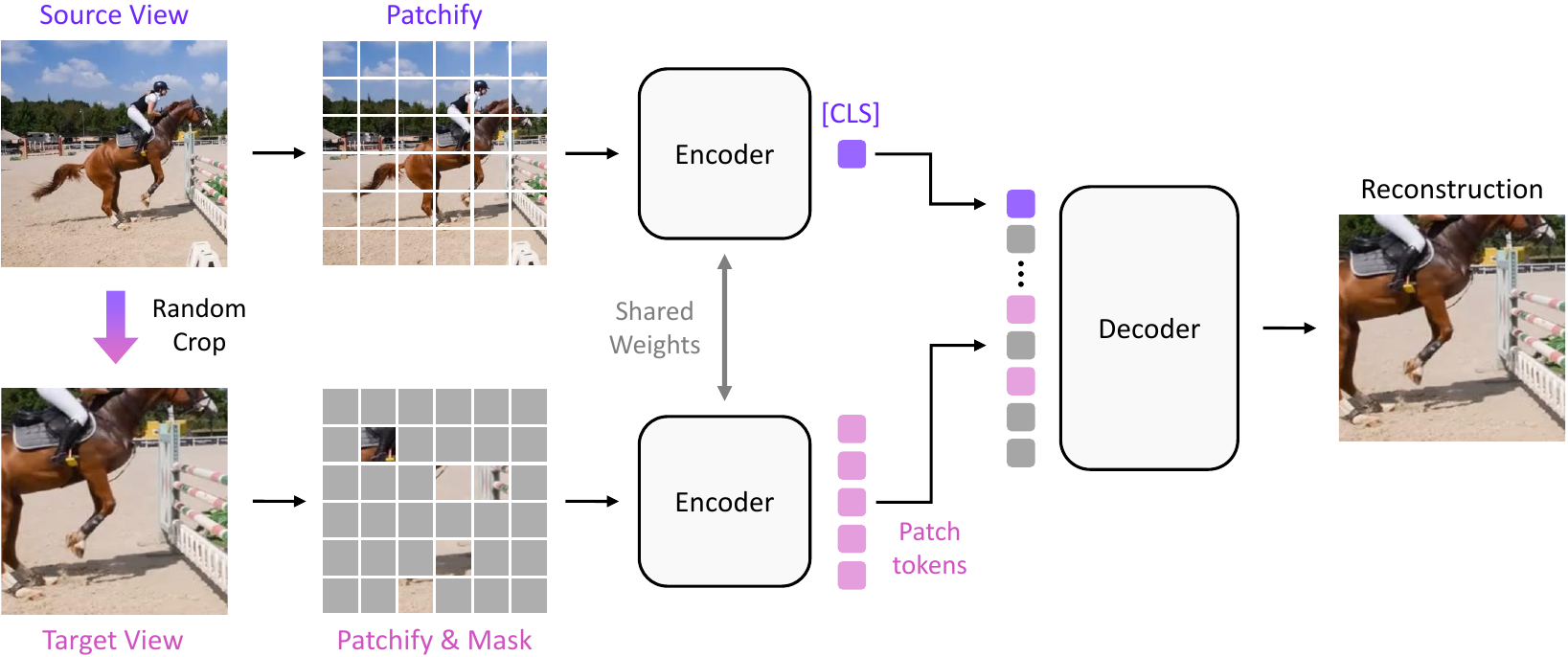}
    
    \caption{
     \textbf{Overview of \sname.} 
     Given a global source view $\mathbf{x}^g$ and a local target view $\mathbf{x}^l$ cropped from $\mathbf{x}^g$, the encoder maps the source view to a single bottleneck token (i.e., [CLS] token) and the target view to a few visible patch tokens under heavy masking (e.g., 90\%). 
     The decoder reconstructs the masked target patches from the visible target patch tokens together with source bottleneck token. Because only a few target patches remain visible, the decoder must rely on the bottleneck token to provide the missing scene context, encouraging it to capture fine-grained \whatiswhere scene composition.}
    \label{fig:framework}
    \vspace{-4mm}
\end{figure*}


\vspace{1mm}
\noindent {\bf Claim.}
Our goal is to learn a compact visual state representation that enables dynamics-aware scene understanding and is transferable to real-world video applications (e.g., robot learning).
To this end, we posit that such a representation must capture \whatiswhere scene composition: 
it should preserve which semantic entities are present in the scene, and how they are semantically arranged within the overall scene composition.
This enables reliable detection of subtle changes and interactions across observations---i.e., understanding \whatmoveswhere in the scene.

\vspace{1mm}
\noindent {\bf Overview.}
We propose \sname, which learns visual states through a global-to-local reconstruction objective~(\cref{fig:framework}).
The name \sname reflects our core design: reconstructing an arbitrary \textbf{\underline{Cro}}pped local view from a \textbf{\underline{Bo}}ttleneck token that summarizes the global scene context of the source view. 
This encourages the visual state representation to capture \whatiswhere in the scene.

\vspace{2mm}
\noindent {\bf Input views.}
Given a video, we sample a frame and construct a pair of views: 
a global source view $\mathbf{x}^g$, obtained by a global crop, and a local target view $\mathbf{x}^l$, by further cropping $\mathbf{x}^g$~\cite{eymael2024efficient}.
We patchify each view into $N$ non-overlapping patches, denoted by $\{\mathbf{x}_i^g\}_{i=1}^N$ and $\{\mathbf{x}_i^l\}_{i=1}^N$.
Note that the source view contains all the information in the target view, as $\mathbf{x}^l$ is spatially contained within $\mathbf{x}^g$.

\vspace{2mm}
\noindent {\bf Siamese encoder and Masking.}
We encode both the source and target views using a shared-weight Siamese encoder~\cite{gupta2023siamese,bromley1993signature}.
The source view $\mathbf{x}^g$ is processed without masking, whereas the target view $\mathbf{x}^l$ is masked with a high masking ratio $r$ (e.g., 90\%) before being fed into the encoder.
We intentionally apply a higher masking ratio than MAE~\cite{he2022masked} ($75\%$) to prevent the target view from being reconstructed solely from the visible tokens.
Let $\mathcal{M} \subset \{1,\dots,N\}$ denote the set of masked patch indices, where $|\mathcal{M}|=\lfloor rN \rfloor$, and let $\mathcal{M}^c=\{1,\dots,N\}\setminus\mathcal{M}$ denote the set of visible patch indices. The source and target views are then encoded as
\begin{align} \label{eq:encode}
    \mathbf{z}_{\mathrm{cls}}^g,\; \{\mathbf{z}_i^g\}_{i=1}^{N} 
    &= \mathrm{Encoder}\!\left(\{\mathbf{x}_i^g\}_{i=1}^{N}\right), \\
    \mathbf{z}_{\mathrm{cls}}^l,\; \{\mathbf{z}_i^l\}_{i\in\mathcal{M}^c} 
    &= \mathrm{Encoder}\!\left(\{\mathbf{x}_i^l\}_{i\in\mathcal{M}^c}\right).
\end{align}
Here, $\mathbf{z}_{\mathrm{cls}}^g$ and $\mathbf{z}_{\mathrm{cls}}^l$ denote the [CLS] tokens of the source and target views, respectively, while $\{\mathbf{z}_i^g\}_{i=1}^{N}$ and $\{\mathbf{z}_i^l\}_{i\in\mathcal{M}^c}$ denote the corresponding patch tokens.

\vspace{2mm}
\noindent {\bf Decoder.}
We reconstruct the masked target view using a Transformer decoder consisting of stacked self-attention and MLP blocks.
Specifically, we first restore the full target token sequence by inserting a learnable mask token at the masked positions and adding positional embeddings. 
The source [CLS] token $z^g_{cls}$, which serves as a single bottleneck token for the source view, is then concatenated with the restored target patch tokens and fed into the decoder.
The masked target patches are reconstructed as
\begin{equation}
\{\hat{\mathbf{x}}_i^l\}_{i=1}^{N}
=
\mathrm{Decoder}\!\left(
\left[\mathbf{z}_{\mathrm{cls}}^g;\, \{\tilde{\mathbf{z}}_i^l\}_{i=1}^{N}\right]
\right),
\end{equation}
where $\{\tilde{\mathbf{z}}_i^l\}_{i=1}^{N}$ is the restored target token sequence.
In this design, the decoder must infer the missing target content from two complementary signals: sparse local hints from the visible target patches and global scene context carried by the bottleneck token $\mathbf{z}_{\mathrm{cls}}^g$. As a result, $\mathbf{z}_{\mathrm{cls}}^g$ is encouraged to preserve the information not only which semantic objects are present in the scene but also where each of them is located, so that the masked target region can be reconstructed faithfully.


\vspace{2mm}
\noindent {\bf Objective.}
We train the encoder and decoder jointly by minimizing the mean squared error over the masked target patches, using normalized pixel targets~\cite{he2022masked}. 
The loss is defined as
\begin{equation}
\mathcal{L}
=
\frac{1}{|\mathcal{M}|}
\sum_{i\in\mathcal{M}}
\left\|
\hat{\mathbf{x}}_i^l - \mathbf{x}_i^l
\right\|_2^2.
\end{equation}

\vspace{2mm}
\noindent {\bf Why train on video datasets?}
\sname constructs both views from a single frame and therefore does not require multiple video frames for pre-training. 
We nevertheless train \sname on a video dataset to ensure fair comparison with prior works~\cite{jang2024visual,kim2025token,gupta2023siamese,eymael2024efficient}.

\begin{table*}[t]
\centering
\small
\setlength{\tabcolsep}{4pt}
\renewcommand{\arraystretch}{1.07}
\caption{
{\bf Experimental results on vision-based robot policy learning.}
We evaluate imitation learning performance on Franka Kitchen~\cite{gupta2019relay}
and DeepMind Control Suite (DMC)~\cite{tassa2018deepmind}. 
The agents are trained on frozen visual representations extracted from a ViT-S/16 encoder pre-trained on the Kinetics-400~\cite{kay2017kinetics}, except for DINOv2.
Success rates (\%) are reported for Franka Kitchen and normalized scores are reported for
DMC.
The best results are \textbf{bold}, second-best are \underline{underlined}.
\sdag~denotes numbers taken from~\cite{jang2024visual}.
We report the gains of our method over the previous state-of-the-art method.
}

\label{tab:franka_cortex_combined}

\resizebox{1\linewidth}{!}{
\begin{tabular}{lccccccccc}
\toprule

& \multicolumn{5}{c}{\textbf{Franka Kitchen}}
& \multicolumn{4}{c}{\textbf{DeepMind Control Suite}} \\

\cmidrule(r{6pt}){2-6}
\cmidrule(l{6pt}){7-10}

Model
& \phantom{x}Knob on\phantom{x}
& \phantom{x}Light on\phantom{x}
& Sdoor open
& Ldoor open
& Micro open
& walker/stand
& walker/walk
& reacher/easy 
& finger/spin \\

\midrule

\dagn{MAE}
& 12.0{\scriptsize$\pm$3.3}
& 24.3{\scriptsize$\pm$4.2}
& 71.5{\scriptsize$\pm$4.3}
& 12.8{\scriptsize$\pm$3.9}
& 10.0{\scriptsize$\pm$2.8}
& -
& -
& -
& - \\

\dagn{DINO}
& 27.0{\scriptsize$\pm$3.2}
& 44.3{\scriptsize$\pm$6.5}
& 77.0{\scriptsize$\pm$5.0}
& 16.5{\scriptsize$\pm$2.5}
& 28.5{\scriptsize$\pm$4.8}
& -
& -
& -
& - \\

DINOv2
& 25.0{\scriptsize$\pm$1.4}
& 46.6{\scriptsize$\pm$5.8}
& 87.8{\scriptsize$\pm$2.7}
& 17.6{\scriptsize$\pm$2.4}
& 21.8{\scriptsize$\pm$3.5}
& 81.2{\scriptsize$\pm$3.7}
& 38.2{\scriptsize$\pm$2.1}
& \underline{87.6}{\scriptsize$\pm$3.8}
& 67.4{\scriptsize$\pm$0.8} \\

\midrule

\dagn{SiamMAE}
& 16.8{\scriptsize$\pm$4.4}
& 36.5{\scriptsize$\pm$7.0}
& 68.0{\scriptsize$\pm$7.9}
& 17.3{\scriptsize$\pm$3.7}
& 13.5{\scriptsize$\pm$4.8}
& -
& -
& -
& - \\

CropMAE
& 33.0{\scriptsize$\pm$3.1}
& 65.0{\scriptsize$\pm$5.1}
& 89.6{\scriptsize$\pm$2.1}
& 22.6{\scriptsize$\pm$2.2}
& 25.0{\scriptsize$\pm$2.9}
& 59.9{\scriptsize$\pm$4.5}
& 33.8{\scriptsize$\pm$3.7}
& 75.4{\scriptsize$\pm$2.8}
& \textbf{69.9}{\scriptsize$\pm$0.8} \\

RSP
& 33.0{\scriptsize$\pm$2.5}
& 44.8{\scriptsize$\pm$9.3}
& 89.6{\scriptsize$\pm$3.8}
& 25.8{\scriptsize$\pm$4.1}
& 33.4{\scriptsize$\pm$1.7}
& 77.3{\scriptsize$\pm$2.5}
& 64.6{\scriptsize$\pm$5.0}
& 63.1{\scriptsize$\pm$3.6}
& 68.5{\scriptsize$\pm$0.7} \\

ToBo
& \underline{58.4}{\scriptsize$\pm$5.1}
& \underline{80.6}{\scriptsize$\pm$4.8}
& \underline{98.4}{\scriptsize$\pm$1.1}
& \textbf{44.2}{\scriptsize$\pm$10.0}
& \underline{51.2}{\scriptsize$\pm$6.1}
& \underline{87.0}{\scriptsize$\pm$3.0}
& \underline{77.7}{\scriptsize$\pm$4.4}
& 87.5{\scriptsize$\pm$3.3}
& \underline{69.6}{\scriptsize$\pm$0.8} \\

\midrule

\sname (Ours)
& \textbf{65.6}{\scriptsize$\pm$3.7}
& \textbf{87.6}{\scriptsize$\pm$2.2}
& \textbf{99.4}{\scriptsize$\pm$1.0}
& \underline{41.2}{\scriptsize$\pm$5.6}
& \textbf{64.8}{\scriptsize$\pm$4.4}
& \textbf{92.0}{\scriptsize$\pm$1.4}
& \textbf{80.8}{\scriptsize$\pm$3.0}
& \textbf{95.8}{\scriptsize$\pm$2.1}
& \textbf{69.9}{\scriptsize$\pm$0.7} \\

\gain{$\Delta$~{prev. SOTA}}
& \gain{+7.2}
& \gain{+7.0}
& \gain{+1.0}
& \lose{-3.0}
& \gain{+13.6}
& \gain{+5.0}
& \gain{+3.1}
& \gain{+8.3}
& \gain{+0.0} \\



\bottomrule
\end{tabular}
}
\label{tab:robot}
\end{table*}

\section{Experiments}
\label{sec:experiments}

We first demonstrate the effectiveness of \sname on vision-based robot policy learning benchmarks, including robotic manipulation and locomotion~(\cref{sec:robot_learning}).
We then analyze the learned representation to better understand its properties.
{\bf 1)} Through reconstruction visualizations, we examine how well the state token captures \whatiswhere scene composition~(\cref{sec:reconstruction}).
{\bf 2)} Through perceptual straightness in video, we analyze how this representation supports \whatmoveswhere understanding across observations~(\cref{sec:perceptual}).

\subsection{Implementation details}
\label{sec:implementation_details}

\noindent {\bf Pre-training.}
For a fair comparison, we follow the pre-training setup of RSP~\cite{jang2024visual}.
We use ViT-S/16~\cite{dosovitskiyimage} and pre-train on the Kinetics-400 dataset~\cite{kay2017kinetics} for 400 epochs with a repeated sampling value of 2~\cite{hoffer2020augment,feichtenhofer2022masked}.
For spatial views $\mathrm{x}^g$ and $\mathrm{x}^l$, we adopt the global-to-local cropping configuration of \cite{eymael2024efficient}, with global and local crop scales of [0.5, 1.0] and [0.3, 0.6], respectively. 
We apply a masking ratio of 90\% to the target view $\mathrm{x}^l$.
The decoder consists of 8 layers with an embedding dimension of 512.
We optimize the model using AdamW~\cite{loshchilov2017decoupled} with a batch size of 1536.
See Appendix.~\ref{appx:imple_details} for full implementation details.

\vspace{1mm}
\noindent {\bf Competitors.} 
We compare our method with standard SSL methods (MAE~\cite{he2022masked}, DINO~\cite{caron2021emerging}, and DINOv2~\cite{oquab2023dinov2}) and dynamic-scene SSL methods (SiamMAE~\cite{gupta2023siamese}, CropMAE~\cite{eymael2024efficient}, RSP~\cite{jang2024visual}, and ToBo~\cite{kim2025token}).

\subsection{Vision-based Robot Learning}
\label{sec:robot_learning}

In this section, we evaluate our method on two vision-based policy learning benchmarks, Franka Kitchen and the DeepMind Control Suite (DMC), covering robotic manipulation and locomotion tasks in simulated environments.

\vspace{1mm}
\noindent {\bf Evaluation setup.}
Across both benchmarks, we freeze the pre-trained visual backbone and train an MLP policy head via behavior cloning using a mean squared error (MSE) loss. A batch normalization layer is applied to the policy input, and results are reported with 95\% confidence intervals over multiple independent runs.

\vspace{1mm}
\noindent {\bf Franka Kitchen.}
Following the experimental protocol of RSP~\cite{jang2024visual}, we evaluate our method and competing baselines on five tasks from the Franka Kitchen benchmark~\cite{gupta2019relay}.
The policy input is formed by concatenating the visual embedding with the robot proprioceptive observations, and the policy head is implemented as an MLP with two hidden layers of 256 units each. Visual observations are captured from left and right camera viewpoints as $224 \times 224$ RGB images.
The policy is trained with a batch size of 32 and 25 expert demonstrations for 20{,}000 gradient steps, with online evaluation in simulation performed every 1{,}000 training steps. We report the mean peak success rate, averaged over 10 independent runs across five random seeds and two camera viewpoints.

\vspace{1mm}
\noindent {\bf DeepMind Control Suite.}
We evaluate our method and competing baselines on four tasks from the DeepMind Control Suite (DMC)~\cite{tassa2018deepmind}, spanning both manipulation and locomotion tasks.
The policy input is formed solely from the visual embedding, without proprioceptive observations, and the policy head is implemented as an MLP with three hidden layers of 256 units each. Visual observations are captured as $224 \times 224$ RGB images with a history window of three consecutive frames.
The policy is trained with a batch size of 256 and 100 expert demonstrations for 100 epochs, with online evaluation performed every 5 epochs. We report the mean peak normalized score, averaged over 10 independent runs across 10 random seeds.

\begin{table*}[t]
\centering
\small
\renewcommand{\arraystretch}{1.07}
\caption{
{\bf Scalability of our method.} 
Success rates (\%) are reported for ViT-S/16, ViT-B/16, and ViT-L/16 backbones pre-trained on Kinetics-400~\cite{kay2017kinetics} for 100 epochs. While ViT-S results are provided for \sname, ViT-B and ViT-L scales include comparisons with SiamMAE~\cite{gupta2023siamese}, RSP~\cite{jang2024visual}, and ToBo~\cite{kim2025token}. $\Delta$ denotes the performance margin relative to the best-performing baseline in each scale. Best results are in \textbf{bold} and second best are \underline{underlined}. \sname results are averaged over 10 trials, while baseline results and those marked with $^\dagger$ are from \cite{kim2025token}.}
\label{tab:scalibility}
\begin{tabular}{llccccc|c}
\toprule
Arch. & Method & \phantom{x}Knob1 on\phantom{x} & \phantom{x}Light on\phantom{x} & Sdoor open & Ldoor open 
& Micro open 
& 5-task Avg. \\
\midrule

\rowcolor{gray!15} {ViT-S/16}
& CroBo (Ours)  
& {57.6}{\scriptsize$\pm$8.1} 
& {81.6}{\scriptsize$\pm$2.1} 
& {98.6}{\scriptsize$\pm$1.4} 
& {36.8}{\scriptsize$\pm$3.8} 
& {50.4}{\scriptsize$\pm$5.8}
& {65.0}{\scriptsize$\pm$4.2} \\

\midrule

\multirow{5}{*}{ViT-B/16}
& SiamMAE$^\dagger$
& 18.0{\scriptsize$\pm$2.0} 
& 34.0{\scriptsize$\pm$2.0} 
& 80.7{\scriptsize$\pm$3.1} 
& 18.7{\scriptsize$\pm$1.2} 
& 19.3{\scriptsize$\pm$6.1}
& 34.1{\scriptsize$\pm$2.9} \\

& RSP$^\dagger$
& 24.7{\scriptsize$\pm$3.1} 
& 51.7{\scriptsize$\pm$9.1}
& 87.3{\scriptsize$\pm$2.3}
& 23.3{\scriptsize$\pm$7.6} 
& 26.7{\scriptsize$\pm$2.3}
& 42.7{\scriptsize$\pm$4.9} \\

& ToBo$^\dagger$
& \underline{46.7}{\scriptsize$\pm$6.4}
& \underline{78.7}{\scriptsize$\pm$7.6}
& \underline{95.3}{\scriptsize$\pm$1.2}
& \textbf{47.3}{\scriptsize$\pm$5.0}
& \underline{37.3}{\scriptsize$\pm$4.6}
& \underline{61.1}{\scriptsize$\pm$5.0} \\

& CroBo (Ours) 
& \textbf{64.4}{\scriptsize$\pm$2.4} 
& \textbf{86.2}{\scriptsize$\pm$3.7}
& \textbf{98.0}{\scriptsize$\pm$1.0} 
& \underline{41.0}{\scriptsize$\pm$6.6} 
& \textbf{62.8}{\scriptsize$\pm$3.4}
& \textbf{70.5}{\scriptsize$\pm$3.4} \\

\cmidrule(lr){2-8}
& \gain{$\Delta$~{prev. SOTA}}
& \gain{+17.7} 
& \gain{+7.5} 
& \gain{+2.7} 
& \lose{-6.3} 
& \gain{+25.5}
& \gain{+9.4} \\

\midrule

\multirow{5}{*}{ViT-L/16}
& SiamMAE$^\dagger$
& 20.7{\scriptsize$\pm$3.1}
& 34.0{\scriptsize$\pm$4.0} 
& 76.0{\scriptsize$\pm$2.0} 
& 12.7{\scriptsize$\pm$6.4} 
& 22.0{\scriptsize$\pm$0.0}
& 33.1{\scriptsize$\pm$3.1} \\

& RSP$^\dagger$
& 26.7{\scriptsize$\pm$2.3}
& 48.0{\scriptsize$\pm$2.0}
& 88.0{\scriptsize$\pm$2.0}
& 22.7{\scriptsize$\pm$8.3}
& 23.3{\scriptsize$\pm$4.2}
& 41.7{\scriptsize$\pm$3.8} \\

& ToBo$^\dagger$
& \underline{54.7}{\scriptsize$\pm$5.0}
& \underline{75.3}{\scriptsize$\pm$4.2}
& \underline{94.0}{\scriptsize$\pm$3.5}
& \textbf{50.0}{\scriptsize$\pm$2.0} 
& \underline{42.7}{\scriptsize$\pm$6.1}
& \underline{63.3}{\scriptsize$\pm$4.2} \\

& CroBo (Ours)
& \textbf{62.4}{\scriptsize$\pm$3.1} 
& \textbf{85.8}{\scriptsize$\pm$3.7} 
& \textbf{98.6}{\scriptsize$\pm$1.0}
& \underline{42.8}{\scriptsize$\pm$5.3}
& \textbf{66.0}{\scriptsize$\pm$5.0}
& \textbf{71.1}{\scriptsize$\pm$3.6} \\

\cmidrule(lr){2-8}
& \gain{$\Delta$~{prev. SOTA}}
& \gain{+7.7} 
& \gain{+10.5} 
& \gain{+4.6} 
& \lose{-7.2} 
& \gain{+23.3}
& \gain{+7.8} \\

\bottomrule
\end{tabular}
\end{table*}
\vspace{1mm}
\noindent {\bf Reproducibility.}
Benchmark performance in vision-based robot learning can exhibit substantial variance across runs due to random seed sensitivity and environment version discrepancies.
While prior work commonly reports results averaged over 5 trials (e.g., RSP~\cite{jang2024visual}, ToBo~\cite{kim2025token}), we evaluate each method over 10 trials within a unified environment to facilitate more reliable comparisons.
For methods with publicly available checkpoints (DINOv2~\cite{oquab2023dinov2}, CropMAE~\cite{eymael2024efficient}, RSP~\cite{jang2024visual}, and ToBo~\cite{kim2025token}), we re-evaluate directly using the released weights.
For MAE~\cite{he2022masked}, DINO~\cite{caron2021emerging}, and SiamMAE~\cite{gupta2023siamese}, whose Kinetics-400~\cite{kay2017kinetics} pretrained checkpoints are not publicly available, we adopt the results reported in RSP~\cite{jang2024visual}.

\vspace{1mm}
\noindent {\bf Result.} 
\cref{tab:franka_cortex_combined} reports the performance of our method and prior self-supervised learning methods on the Franka Kitchen~\cite{gupta2019relay} and the DeepMind Control Suite benchmarks~\cite{tassa2018deepmind}. Overall, \sname consistently outperforms existing approaches across most tasks.
On the Franka Kitchen benchmark, \sname achieves the best performance on four out of five tasks, substantially improving over the previous state-of-the-art. In particular, it yields large gains on Micro open (\textbf{+13.6\%}), Knob on (\textbf{+7.2\%}), and Light on (\textbf{+7.0\%}).
On the DeepMind Control Suite, our method similarly establishes new best results across several tasks, with the most notable improvements on reacher/easy (\textbf{+8.3\%}), walker/stand (\textbf{+5.0\%}), and walker/walk (\textbf{+3.1\%}).
Importantly, these improvements are observed across both robotic manipulation and locomotion tasks. This suggests that the representations learned by \sname capture visual features that generalize across diverse embodied control problems, rather than being tailored to a specific domain.

\vspace{1mm}
\noindent {\bf Scaling behavior.} 
To evaluate scalability across model capacities, we benchmark CroBo against baselines using ViT-B/16 and ViT-L/16 backbones.
All models are pre-trained on Kinetics-400~\cite{kay2017kinetics} for 100 epochs and evaluated on Franka Kitchen~\cite{gupta2019relay}.
As shown in \cref{tab:scalibility}, \sname consistently outperforms SiamMAE~\cite{gupta2023siamese}, RSP~\cite{jang2024visual}, and ToBo~\cite{kim2025token} across all architecture scales.
The base and large variants achieve 5-task average success rates of 70.5\% and 71.1\%, respectively, surpassing the prior state of the art by substantial margins of \textbf{+9.4\%} and \textbf{+7.8\%}.
Remarkably, even our smallest backbone (ViT-S/16) achieves a 65.0\% average success rate, outperforming all baselines built on the much larger ViT-L/16.
This shows that the gains of \sname arise not from model scale, but from a stronger representation that generalizes across architectures of varying capacity.

\subsection{Qualitative Analysis}
\label{sec:reconstruction}
\begin{figure*}[!t]
\centering
\includegraphics[width=0.975\linewidth]{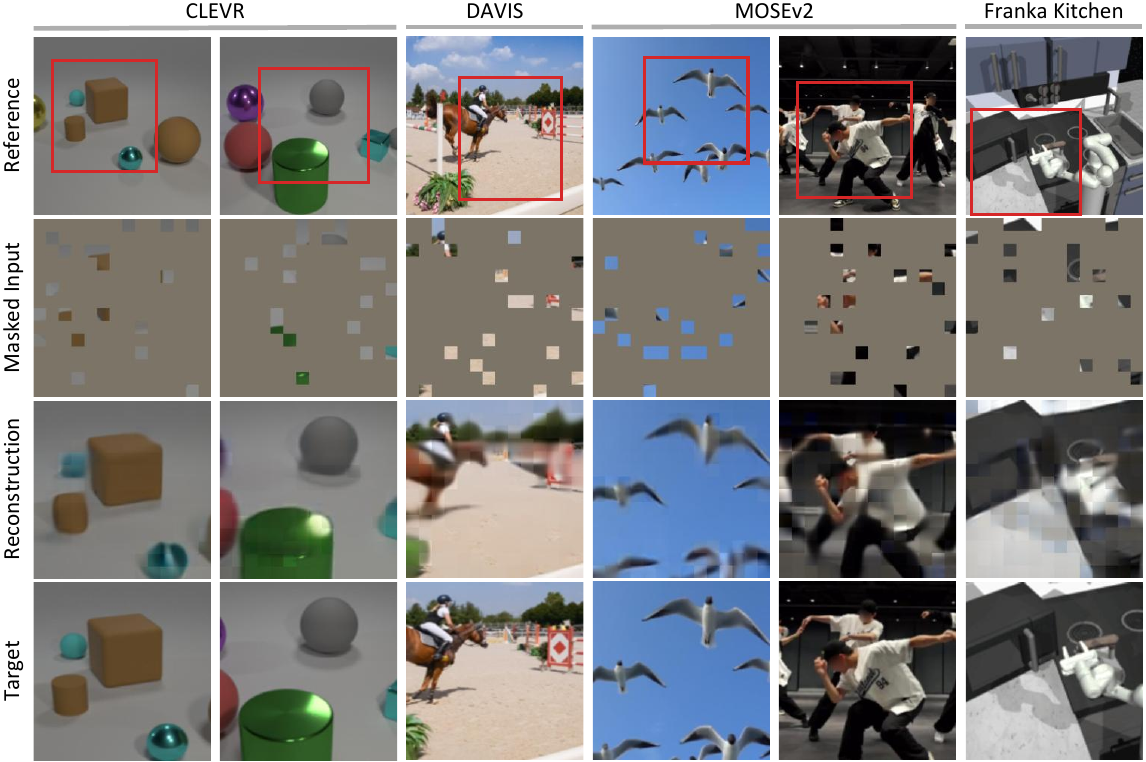}
\caption{
    {\bf Reconstruction of \sname.} 
    We visualize image reconstructions on CLEVR~\cite{johnson2017clevr}, DAVIS~\cite{pont20172017}, MOSEv2~\cite{ding2025mosev2} and Franka Kitchen~\cite{gupta2019relay}. 
    Using the bottleneck token of the reference view as context, \sname reconstructs a highly masked (90\%) target view cropped from the reference view. 
    The results show that the bottleneck token from the reference view provides sufficient information to recover overall scene structure, including object identity, location, and spatial relationships.
    (e.g., in column 1, the two cyan spheres are not visible in the masked input, yet their positions are correctly reconstructed using the reference bottleneck token).
} 
\vspace{-3mm}
\label{fig:recon}
\end{figure*}

In this section, we examine how well the bottleneck token captures \whatiswhere scene composition through reconstruction visualizations.
We follow the same reconstruction setup used during training~(\cref{fig:recon}).

\vspace{1mm}
\noindent {\bf Target datasets.} We visualize image reconstructions on CLEVR~\cite{johnson2017clevr}, DAVIS~\cite{pont20172017}, MOSEv2~\cite{ding2025mosev2}, and Franka Kitchen~\cite{gupta2019relay}.
CLEVR provides synthetic scenes with simple layouts and well-defined object attributes (e.g., color, shape, and materials), enabling evaluation of \whatiswhere scene composition.
DAVIS considers natural video scenes with a dominant object, whereas MOSEv2 features more complex and crowded dynamic scenes with multiple interacting instances.
Franka Kitchen provides complex robotic manipulation scenes.

\noindent {\bf Analysis.}
The first column shows that the bottleneck token preserves object attributes along with spatial locations: the two cyan spheres are accurately reconstructed, even though they are fully occluded in the masked input.
The second column further demonstrates that the representation retains fine-grained object and scene details, including metallic reflectance and shadows.
The third through sixth columns show that these properties remain robust in more complex and cluttered scenes.
In the third column, CroBo successfully recovers the horse's overall shape and the rider's pose from only a few visible human parts.
Moreover, the fourth through sixth columns illustrate that the model faithfully recovers multiple scene elements and their complex configurations.
These results suggest that the bottleneck token encodes a coherent scene representation with a strong understanding of semantic arrangement even under increased complexity.
This capability contributes to the strong robotic performance of CroBo by capturing reliable \whatiswhere scene composition during manipulation.
Additional results are provided in Appendix~\ref{appx:recon}.


\subsection{Perceptual Straightness in Video}
\label{sec:perceptual}
In this section, we analyze the temporal geometry of representations of our dynamics-aware SSL model. 
Specifically, we investigate whether representations evolve smoothly over time when processing natural videos using the concept of \emph{perceptual straightening} \cite{henaff2019perceptual, henaff2021primary}.
%

\begin{figure*}[t]
\centering
\includegraphics[width=1\linewidth]{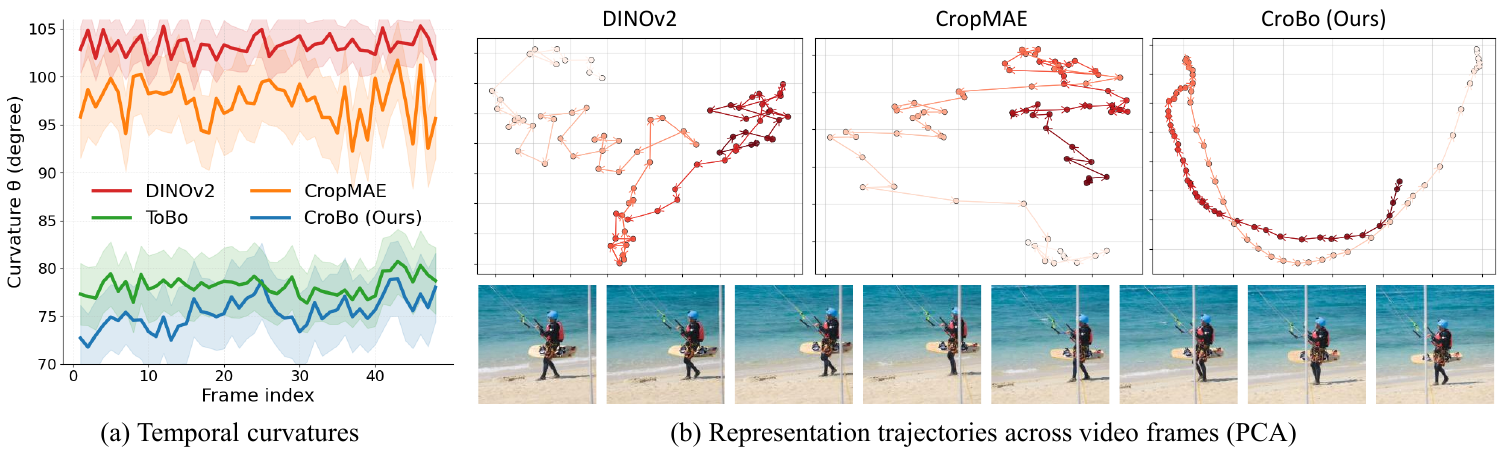}
\caption{
    {\bf Perceptual straightness of representation dynamics in video.}
    (a) Local curvature of representation trajectories measured on DAVIS videos. Curvature is computed as the angle between consecutive representation differences, averaged across videos. Lower curvature indicates smoother temporal dynamics. CroBo consistently exhibits lower curvature than prior models such as DINOv2 and CropMAE, suggesting more temporally coherent representations that better preserve \whatmoveswhere across frames.
    (b) Representation trajectories across video frames visualized using PCA. Each point corresponds to the representation of a frame, and colors indicate temporal progression.
    %
    CroBo produces smooth and locally linear trajectories that follow the natural evolution of the scene compared to other models.
}
\vspace{-3mm}
\label{fig:perceptual}
\end{figure*}

\vspace{1mm}
\noindent {\bf Perceptual straightness.}
%
Perceptual straightness is the property of perceptual representations in which the visual system transforms inputs to follow a straight path in the representation space. 
Since observations vary in a highly nonlinear way in the pixel space, it is infeasible to make predictions from raw observations \cite{niu2024learning}.
Therefore effective video understanding systems \cite{henaff2019perceptual, bagad2025chirality} compress raw observations into a visual state that has a locally straight temporal trajectory,
enabling future state prediction via extrapolation.

\vspace{1mm}
\noindent {\bf Measuring perceptual straightness.}
%
Local curvature is the main metric to assess the perceptual straightness of representations \cite{henaff2019perceptual, interno2025ai}.
%
Let $z_t$ denote the representation of frame $x_t$.
The temporal trajectory of a video in representation space is the sequence $\{z_t\}_{t=1}^T$. 
We measure the local curvature using the angle between consecutive difference vectors $\Delta z_t = z_{t+1} - z_t$ and $\Delta z_{t+1} = z_{t+2} - z_{t+1}$ as 
\begin{equation}
\theta_t
=
\arccos
\left(
\frac{\Delta z_t }{\|\Delta z_t\|_2 }
\cdot
\frac{\Delta z_{t+1} }{\|\Delta z_{t+1}\|_2 }
\right).
\end{equation}
Lower curvature indicates that representation trajectories evolve more linearly over time, suggesting temporally consistent features that effectively encode \whatmoveswhere.
In other words, a model understanding smooth motion will produce a smooth trajectory of representations \cite{henaff2019perceptual}.
%


\vspace{1mm}
\noindent {\bf Quantitative comparison.}
We measure local curvature on videos of the DAVIS validation set and compare across models.
For each video containing at least 50 frames, we compute the trajectory and the local curvature for the first 50 frames and plot the average local curvature at \cref{fig:perceptual}(a). 
CroBo achieves an average curvature of $75.4^\circ$, compared to $103.28^\circ$ of DINOv2.
%
The lower curvature signal that representations of our model follow a locally linear path.
%
This suggests that \sname is able to capture subtle temporal differences between adjacent frames and maintain a coherent representation of \whatmoveswhere.
%
%

\vspace{1mm}
\noindent {\bf An example trajectory.}
%
%
%
As an intuitive example, we evaluate representations of frames from the kite-walk video in DAVIS~\cite{pont20172017}, compute two principal components of the trajectory via PCA, and plot it at \cref{fig:perceptual}(b) following \cite{bagad2025chirality}. 
While DINOv2 and CropMAE generated highly jagged trajectories, \sname produced a smooth trajectory,
%
compliant to the contents of the video.
While the person moves right and then left in the video, the representation also moves back and forth in the first principal component. Additional results are provided in Appendix.~\ref{appx:perceptual}.

{}

\subsection{Ablation Studies}
    \label{sec:ablation}

\subsubsection{Ablation on Target View Construction}
    \label{sec:ablation_target_view}
    {\iftrue
        In this section, we compare our {\it global-to-local} formulation with alternative {\it source-to-target} relationships for visual state representation learning (SRL).
        Following SiamMAE~\cite{gupta2023siamese}, we consider a {\it current-to-future} relationship in which two frames are sampled from a video; we refer to this formulation as \textbf{Time}.
        In contrast, our method adopts a {\it global-to-local} relationship~\cite{eymael2024efficient}, where the target view is obtained by randomly cropping the source view; we refer to this formulation as \textbf{Crop}.
        We also consider a hybrid variant, \textbf{Time + Crop}, in which the target is constructed by cropping a temporally subsequent frame.

        \vspace{2mm}
        \noindent{\bf Time vs. Crop.}
        We compare Time and Crop by pre-training ViT-S/16 on Kinetics-400 for 100 epochs and evaluating on Franka Kitchen.
        As shown in \cref{tab:ablation_target_view}, Crop consistently outperforms Time across all five tasks.
        We attribute this result to the quality of the supervisory signal.
        Under Crop, the target is spatially grounded because it is fully contained in the source view, making reconstruction well-defined.
        By contrast, Time introduces ambiguity from object motion, camera motion, and appearance changes across frames.
        These results suggest that learning spatial scene composition through Crop provides a stronger basis for visual state representation than relying on temporal correspondence.
            
        \vspace{2mm}
        \noindent{\bf Time + Crop.}
        One may expect that combining temporal variation with spatial grounding would provide a richer learning signal.
        However, \cref{tab:ablation_target_view} shows that Time+Crop performs worse than both Time and Crop.
        We conjecture that combining the two makes the prediction target less coherent: the model must jointly account for spatial localization and temporal change, which increases target ambiguity and weakens supervision.
        This result further supports our choice of the Crop-based global-to-local formulation. 
    \fi}

\subsubsection{Ablation on Masking ratio}
\label{sec:ablation_masking_ratio}

We vary the masking ratio of the target scene from 75\% to 95\% to verify whether limiting target-view information encourages the decoder to more effectively exploit the visual scene representation encoded from the source scene.
As shown in \cref{tab:ablation_mask_ratio}, performance improves substantially as the masking ratio increases from 75\% to 90\%, supporting our hypothesis that reducing target-side clues encourages the model to make fuller use of the compressed reference information.
Moreover, although our default setting is 90\%, a more aggressive masking ratio of 95\% yields the best performance across all tasks.
These results indicate that further limiting the visible target content encourages the model to draw more effectively on the source scene, leading to the best overall performance.

\newcolumntype{Y}{>{\centering\arraybackslash}X}

\begin{table}[t]
    \centering
    \small
    \renewcommand{\arraystretch}{1.07}
    
    \caption{
        Ablation studies on target view design and masking ratio.
        Models are pre-trained on Kinetics-400 for 100 epochs and evaluated on Franka Kitchen (success rate \%).
        Our method is highlighted in \colorbox{gray!15}{gray}.
    }
    \label{tab:ablation_vertical}
    
    \begin{subtable}[t]{\linewidth}
        \centering
        \begin{tabularx}{\linewidth}{YY|YYYYY}
            \toprule
            Time & Crop & Knob1 & Light & Sdoor & Ldoor & Micro \\
            \midrule
            
            \ding{51} & {}
            & 44.4 & 67.6 & 96.4 & \textbf{36.8} & 49.6 \\
            
            \rowcolor{gray!15}
            {} & \ding{51}
            & \textbf{57.6} & \textbf{81.6} & \textbf{98.6} & \textbf{36.8} & \textbf{50.4} \\
            
            \ding{51} & \ding{51}
            & 38.2 & 62.8 & 90.8 & 22.2 & 28.4 \\
            \bottomrule
        \end{tabularx}
    
        \vspace{0.5mm}
        \caption{\textbf{Target view construction}}
        \label{tab:ablation_target_view}
    \end{subtable}
    
    \vspace{4pt}
    
    \begin{subtable}[t]{\linewidth}
        \centering
        \begin{tabularx}{\linewidth}{YY|YYYYY}
            \toprule
            \multicolumn{2}{c|}{Masking Ratio} & Knob1 & Light & Sdoor & Ldoor & Micro \\
            \midrule
            
            \multicolumn{2}{c|}{75\%}
            & 41.4 & 70.6 & 94.0 & 22.6 & 35.0 \\
            
            \rowcolor{gray!15}
            \multicolumn{2}{c|}{90\%}
            & 57.6 & 81.6 & 98.6 & 36.8 & 50.4 \\
            
            \multicolumn{2}{c|}{95\%}
            & \textbf{59.0} & \textbf{86.6} & \textbf{99.4} & \textbf{41.2} & \textbf{58.0} \\
            
            \bottomrule
        \end{tabularx}
    
        \vspace{0.5mm}
        \caption{\textbf{Masking ratio}}
        \label{tab:ablation_mask_ratio}
    \end{subtable}
    \vspace{-10pt}
\end{table}
\section{Related Work}
\label{sec:related_works}

\noindent \textbf{Self-supervised Learning (SSL)} aims to learn rich and transferable visual representations from unlabeled images or videos.
A large body of SSL methods build on representation alignment across augmented views, using contrastive objectives~\cite{wu2018unsupervised,chen2020simple,he2020momentum,chen2021empirical,caron2020unsupervised} or teacher-student self-distillation~\cite{caron2021emerging,grill2020bootstrap,zhou2021ibot,oquab2023dinov2} to learn invariant features.
Meanwhile, reconstruction-based approaches learn representations by recovering masked pixels or tokens~\cite{bao2021beit,he2022masked,zhou2021ibot,xie2022simmim}.
Despite their success on traditional tasks such as image classification and segmentation, transferable representations for dynamic scene understanding remain underexplored.

\vspace{1mm}
\noindent \textbf{Siamese Masked Autoencoder}~\cite{gupta2023siamese}, 
on the other hand, targets dynamic scene understanding by learning visual correspondence from videos with a Siamese encoder~\cite{bromley1993signature}.
It predicts masked patches in one frame from another, learning patch-wise correspondence across time. 
CropMAE~\cite{eymael2024efficient} shows that similar correspondence can be learned even from static images by reconstructing one cropped view from another.
However, such patch-wise correspondence is suboptimal for downstream video tasks that require a {\it single} compact frame representation, such as robotic manipulation.

\vspace{1mm}
\noindent \textbf{State Representation Learning (SRL)}, introduced by ToBo~\cite{kim2025token}, addresses the above limitation by learning to encode an observation into a single compact visual state (i.e., the bottleneck token).
It learns this state by reconstructing heavily masked target patches from the bottleneck token of a temporally adjacent frame and a few visible target patches.
This encourages temporal relationships to be preserved in the representation, but does not explicitly account for within-frame spatial structure and scene composition.

\vspace{1mm}
\noindent \textbf{Our work} extends this line of work by studying a {\it dynamics-aware} SSL method for learning state representations that capture scene composition.
While we follow prior works in using a Siamese network and paired-view reconstruction~\cite{gupta2023siamese,eymael2024efficient,kim2025token}, our work differs in two key aspects.
First, we explicitly argue that compact visual states for dynamic world understanding should preserve scene composition. 
Second, to achieve this, we propose a novel global-to-local reconstruction framework that learns to encode scene composition into the bottleneck token.





\section{Conclusion}
\label{sec:conclusion}
Effective dynamic scene understanding of video data requires visual state representations that explicitly encode the semantic identities and precise spatial locations of scene elements in a unified manner.
To this end, we introduce \sname, a framework built around a global-to-local reconstruction objective, where a compact bottleneck token reconstructs masked local crops.
Reconstruction analyses show that \sname captures fine-grained \whatiswhere scene composition, and perceptual straightness evaluations further indicate that this spatial grounding enables representations to track \whatmoveswhere across dynamic observations.
Consequently, compact visual representations that preserve pixel-level scene information lead to state-of-the-art performance on vision-based robot policy learning benchmarks.

\section*{Acknowledgements.}
We sincerely thank Stella Yu, Chanyong Lee, Hoseong Kim, and Eunjin Koh for their constructive discussions
and support. We also appreciate Junwha Hong, Sol Park and Minkyu Song for providing insightful feedback.
This work was supported by the Agency For Defense Development Grant Funded by the Korean
Government (912A45701).

{
    \newpage
    \small

}

\clearpage
\setcounter{page}{1}
\appendix
\maketitlesupplementary

\section*{Appendix}
In the supplemental material, we provide:
\vspace{0.3em}

\noindent\hspace*{0.7em} \hyperref[appx:imple_details]{A. Additional Implementation Details}

\noindent\hspace*{0.7em} \hyperref[appx:rel_works]{B. Additional Related Works}

\noindent\hspace*{0.7em} \hyperref[appx:recon]{C. Additional Reconstruction Visualization}

\noindent\hspace*{0.7em} \hyperref[appx:perceptual]{D. Additional Perceptual Straightness Results}


\section{Additional Implementation Details}
\label{appx:imple_details}
\noindent {\bf Decoder.}
Following ToBo~\cite{kim2025token}, we employ a decoder composed of self-attention and multi-layer perceptron (MLP) layers. 
The decoder configuration follows MAE~\cite{he2022masked}, with the detailed settings summarized in \cref{tab:decoder_config}.

\vspace{1.5mm}
\noindent {\bf Pre-training.}
The pre-training details are provided in \cref{tab:pretraining_config}. 
Unless otherwise specified, we follow the setup of RSP~\cite{jang2024visual} for a fair comparison.
For the main comparison, we report pre-training for 400 epochs; with repeated sampling~\cite{hoffer2020augment,gupta2023siamese} of 2, this corresponds to running 200 training epochs in practice, where each sample is seen twice per epoch. 
For ablation studies, we train for 100 epochs and reduce the number of warmup epochs from 40 to 20.

\vspace{1.5mm}
\noindent {\bf Augmentation.}
The augmentation details are provided in \cref{tab:augment_config}. 
Note that the global crop is sampled from the selected frame, and the local crop is then sampled from the global crop, so that the local crop remains fully contained within the global view.
We use global and local crop scales of [0.5, 1.0] and [0.3, 0.6], respectively, following CropMAE~\cite{eymael2024efficient}.
We also apply horizontal flipping in a synchronized manner across global and local views for improved training stability (i.e., if a global view is flipped, the corresponding local view is flipped as well).



\begin{table}[t]
\centering
\renewcommand{\arraystretch}{1.07}
\caption{{\bf Implementation details.}
We follow MAE~\cite{he2022masked} for the decoder, 
RSP~\cite{kim2025token} for pre-training, 
and CropMAE~\cite{eymael2024efficient} for augmentation, except for synchronized horizontal flipping.
}

\begin{subtable}{\columnwidth}
\centering
\vspace{-1mm}
\small
\begin{tabular}{p{0.45\columnwidth} p{0.4\columnwidth}}
\toprule
\textbf{Parameter} & \textbf{Setting} \\
\midrule
Depth & 8 \\
Embedding dimension & 512 \\
Number of attention heads & 16 \\
MLP ratio & 4.0 \\
\bottomrule
\end{tabular}
\vspace{0.5mm}
\caption{{\bf Decoder configuration.}}
\label{tab:decoder_config}
\end{subtable}

\vspace{3mm}

\begin{subtable}{\columnwidth}
\centering
\vspace{-1mm}
\small
\begin{tabular}{p{0.45\columnwidth} p{0.4\columnwidth}}
\toprule
\textbf{Parameter} & \textbf{Setting} \\
\midrule
Batch size & 1536 \\
Epochs & 400 \\
Warmup epochs~\cite{goyal2017accurate} & 40 \\
LR scheduler & Cosine decay~\cite{loshchilov2016sgdr} \\
Learning rate & $1.5 \times 10^{-4}$ \\
Optimizer & AdamW~\cite{loshchilov2017decoupled} \\
Adam $\beta_1, \beta_2$ & (0.9, 0.95)~\cite{chen2020generative} \\
Weight decay & 0.05 \\
Repeated sampling~\cite{hoffer2020augment} & 2 \\
\bottomrule
\end{tabular}
\vspace{0.5mm}
\caption{{\bf Pre-training configuration.}}
\label{tab:pretraining_config}
\end{subtable}

\vspace{3mm}

\begin{subtable}{\columnwidth}
\centering
\vspace{-1mm}
\small
\begin{tabular}{p{0.45\columnwidth} p{0.4\columnwidth}}
\toprule
\textbf{Parameter} & \textbf{Setting} \\
\midrule
Augmentation type & RandomCrop + HFlip \\
Crop aspect ratio & [3/4, 4/3] \\
Global crop scale & [0.5, 1.0] \\
Local crop scale & [0.3, 0.6] \\
Resize & 224 $\times$ 224 \\
Interpolation & Bicubic \\
HFlip probability & 0.5 (Synchronized) \\
\bottomrule
\end{tabular}
\vspace{0.5mm}
\caption{{\bf Augmentation configuration.}}
\label{tab:augment_config}
\end{subtable}

\vspace{-2mm}
\label{tab:implementation_details}
\end{table}

\section{Additional Related Works}
\label{appx:rel_works}

\noindent {\bf Benchmarking Pixel-level video understanding.}
Pixel-level video understanding has been studied through a variety of benchmarks that evaluate fine-grained spatiotemporal understanding in dynamic scenes, such as video object segmentation (VOS).
Early datasets such as DAVIS~\cite{perazzi2016benchmark,pont20172017} and YouTube-VOS~\cite{xu2018youtube} established standard VOS benchmarks, with high-quality dense annotations in DAVIS and large-scale data diversity in YouTube-VOS.
More recent benchmarks~\cite{ding2023mose,ding2025mosev2,ding2023mevis,ding2025mevis} further increase scene complexity and motion ambiguity.
In particular, MOSE~\cite{ding2023mose} and MOSEv2~\cite{ding2025mosev2} introduce crowded and complex scenes with heavy occlusion, frequent disappearance and reappearance, and multiple same-category instances.
In parallel, MeViS~\cite{ding2023mevis} and MeViSv2~\cite{ding2025mevis} extend this line of work to referring video segmentation based on motion expressions.

While these benchmarks evaluate fine-grained scene understanding with dense pixel-level annotations, our work instead learn such understanding in a compact visual state representation (i.e., [CLS] token). 
Therefore, our method is not directly aligned with standard VOS evaluation protocols. 
Instead, we leverage challenging VOS datasets such as DAVIS and MOSEv2 for reconstruction analysis and perceptual straightness, to assess whether the learned representation captures complex scene dynamics.

\clearpage
\noindent {\bf Visual Representations for Sequential World Modeling.}

\noindent Latent world models~\cite{ha2018world,hafner2019learning,hafner2019dream,hafner2020mastering,hafner2023mastering} aim to learn compact states that support long-horizon prediction and decision-making from high-dimensional observations~\cite{bredis2026next}. 
They rely on visual representations to encode observations into these states~\cite{zhou2024dino,baldassarre2025back,karypidis2024dino}, since these representations provide the primary interface to the external environment.
Accordingly, the representation should both capture fine-grained information from observations and be predictive over time.
In this context, \sname may serve as a suitable visual encoder for sequential world modeling as it encodes what-is-where scene information into a compact state and exhibiting temporal straightness~\cite{henaff2019perceptual,henaff2021primary,niu2024learning}.
This further suggests its potential for predictive modeling in representation space, such as next-embedding prediction framework~\cite{xu2025next,bredis2026next,hu2024drivingworld}.

\vspace{2mm}
\noindent {\bf Locality-aware Image Representation Learning.}
Recent work suggests that capturing localized information, such as objects and regions, helps fine-grained image understanding.
In visual representation learning, this has been explored through dense or local correspondence learning~\cite{xie2021unsupervised,lebailly2023global,wang2021dense,wen2025data}, and in vision-language representation learning through region-level alignment between image regions and regional text descriptions~\cite{choi2025goal,woo2026aligning,jing2024fineclip,xie2025fg}.
Our work also prioritizes preserving localized scene details but focuses on encoding this fine-grained composition into a single compact visual state for dynamic scene understanding.

\section{Additional Reconstruction Visualization}
\label{appx:recon}
\begin{figure*}[!ht]
\centering
\vspace{1mm}
\includegraphics[width=1\linewidth]{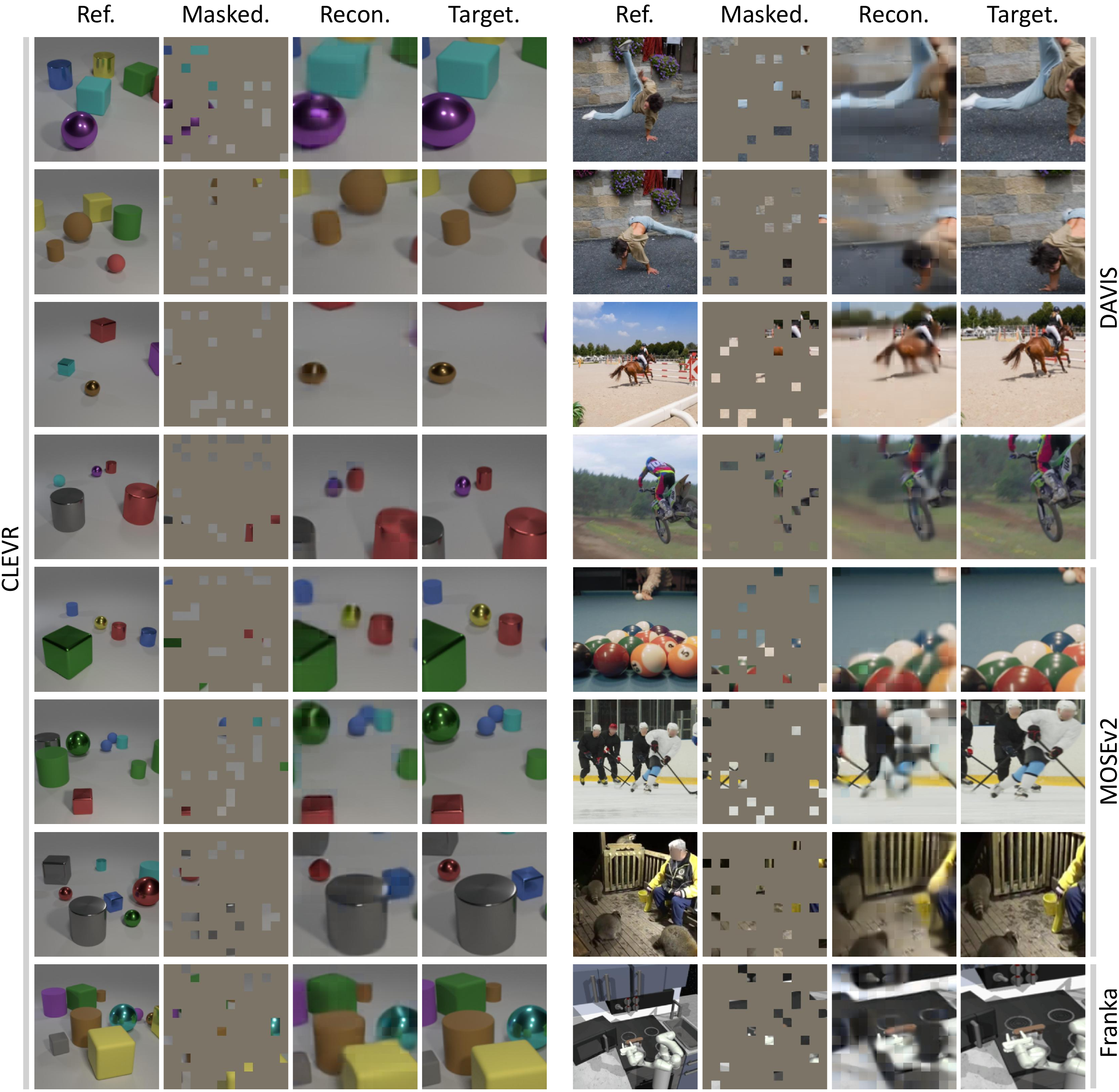}
\caption{
{\bf Reconstruction results of \sname.}
We present image reconstruction examples on CLEVR~\cite{johnson2017clevr}, DAVIS~\cite{pont20172017}, MOSEv2~\cite{ding2025mosev2} and Franka Kitchen~\cite{gupta2019relay}. 
Given the bottleneck token extracted from the reference view as context, \sname reconstructs a heavily masked (90\%) target view spatially cropped from the reference view.
These reconstructions indicate that the bottleneck token captures sufficient information to recover the overall scene structure, including object identities, spatial locations, and their relationships.
}
\label{fig:supple_recon}
\end{figure*}

We provide extensive reconstruction visualizations in ~\cref{fig:supple_recon}, extending the analysis in \cref{sec:reconstruction} to a broader set of cases. Using the bottleneck token from the reference view, \sname reconstructs 90\% masked target views across CLEVR, DAVIS, MOSEv2, and Franka Kitchen.
The results demonstrate that our model faithfully restores object identities and spatial locations.
This consistent performance across diverse scene types further confirms that the bottleneck token effectively captures pixel-level \whatiswhere scene composition.

\section{Additional Perceptual Straightness Results}
\label{appx:perceptual}
\begin{figure*}[!ht]
\centering
\vspace{1mm}
\includegraphics[width=0.97\linewidth]{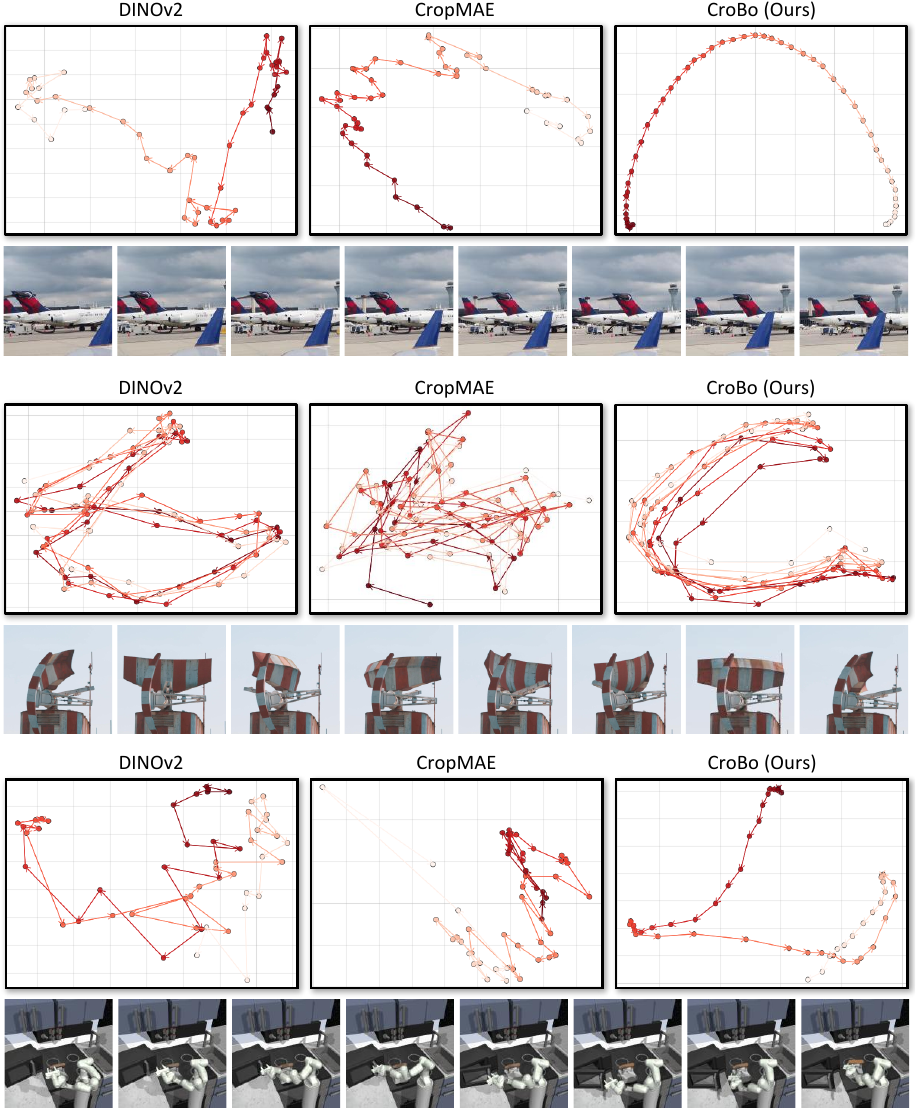}
\caption{
{\bf Perceptual straightness of representation trajectories in video.}
We visualize representation trajectories across video frames using PCA, where each point denotes a frame and colors indicate temporal order.
CroBo produces smooth and locally linear trajectories that follow the natural evolution of the scene, whereas DINOv2 and CropMAE exhibit more irregular and entangled trajectories.
Videos are drawn from MOSEv2 (first and second rows)~\cite{ding2025mosev2} and Franka Kitchen (third row)~\cite{gupta2019relay}.
}
\label{fig:supple_perceptual}
\end{figure*}

To further probe the perceptual straightness of CroBo's learned representations, we present additional qualitative analyses on three representative scenarios in \cref{fig:supple_perceptual}.
We visualize representation trajectories using PCA and compare CroBo with DINOv2~\cite{oquab2023dinov2} and CropMAE~\cite{eymael2024efficient}. 
Together, these examples cover near-linear motion, periodic rotation, and sequential manipulation, illustrating how the learned representation behaves under distinct temporal structures.

\vspace{2mm}
\noindent {\bf Straight taxiing airplane} (MOSEv2, \texttt{4xz71muo}).
This video shows an airplane slowly taxiing along a runway, with smooth camera motion that tracks the airport scene layout. 
The motion is nearly linear from a perceptual standpoint, with minimal viewpoint change and steady progression across the scene.
Our model produces a smooth and nearly linear trajectory that closely follows this temporal progression, indicating that the representation preserves the underlying motion in a consistent and ordered manner.

\vspace{2mm}
\noindent {\bf Rotating radar antenna} (MOSEv2, \texttt{7bh6bqw6}).
This video shows a radar antenna rotating clockwise for five turns, with slight camera motion.
Our model effectively captures this periodic motion and the accompanying appearance changes, producing a coherent and smoothly evolving trajectory that reflects the underlying cyclic structure.
Notably, the trajectory exhibits a repeating C-shaped pattern that matches the Lissajous-like curves obtained when circular motion is projected onto a 2D plane.
DINOv2 also captures a reasonably structured trajectory, whereas CropMAE exhibits more entangled and irregular patterns, suggesting weaker consistency in representing the periodic motion.

\vspace{2mm}
\noindent {\bf Opening microwave} (Franka Kitchen Demo).
This example shows the first two seconds of a Franka Kitchen demo, where the robot arm moves left, grasps the microwave handle, opens the door, and then reaches toward a kettle, while the camera remains fixed.
Our model produces a smooth and interpretable trajectory that follows this sequential manipulation process.
In particular, the turning point at the left L-shaped corner of CroBo’s trajectory corresponds to the moment when the robot hand grasps the microwave handle, suggesting that the learned representation captures a perceptually meaningful transition in the scene.
By contrast, the trajectories of competing methods exhibit irregular, zigzagging patterns with frequent local direction changes, making them less structured and harder to interpret.

\end{document}